\documentclass[conference]{IEEEtran}
\IEEEoverridecommandlockouts

\usepackage[T1]{fontenc}
\usepackage{cite}
\usepackage{amsmath,amssymb}
\usepackage{graphicx}
\usepackage{booktabs}
\usepackage{url}
\usepackage[hidelinks]{hyperref}
\usepackage{balance}
\usepackage{tikz}
\usetikzlibrary{arrows.meta,positioning,fit}
\newcommand{\method}[1]{\textsc{#1}}

\graphicspath{{figures/}}
\begin{document}

\title{A Deployment Study of Identity-Gated\\Drone Gesture Control}

\author{Diyari Mohammed Salih$^{1}$,
        Ilyes Chaabeni$^{1}$,
        and Na\"{\i}ma A\"{\i}t Oufroukh$^{2}$%
\thanks{$^{1}$M2 Smart Aerospace and Autonomous Systems,
Universit\'e Paris-Saclay, \'Evry, France.}%
\thanks{$^{2}$IBISC Laboratory, Universit\'e Paris-Saclay, \'Evry, France.}%
\thanks{E-mails: diyari.m.salih@gmail.com,
chaabeni.ilyes2002@gmail.com,
naima.aitoufroukh@univ-evry.fr.}%
}

\maketitle

\begin{abstract}
Vision-based gesture control accepts commands from any hand in the camera field of view, which is unsafe in shared indoor spaces. 
This paper presents \method{IGate}, an identity-gated control stack that includes gesture control and face tracking, in which commands are admitted only when an enrolled operator is verified. The system performs few-shot user enrolment from 20 initial face frames, without prior user-specific training: verification compares an embedding of the current face crop against the enrolled template by cosine similarity, while face tracking uses proportional correction. Gesture control is achieved by classifying extracted hand landmarks using an RBF-SVM trained on a custom dataset. Additionally, a hierarchical finite-state machine handles mode selection, default, and fallback behaviours.
The approach is tested on a DJI Tello EDU, each component evaluated offline and in-flight across 270 trials (149 flown). Face verification yields a 0.32\% offline equal error rate versus 19.3\% in-flight. Under hover-locked conditions, the RBF-SVM gesture classifier outperforms the geometric rule (0.850 vs. 0.651 accuracy), with 82\% of this gap stemming from the depth channel. All logs and reproduction scripts will be released.
\end{abstract}


\section{Introduction}

Small indoor drones are increasingly operated through vision-based gesture
interfaces, which remove the physical controller and lower the barrier for
novice users.
A camera-based recogniser, however, responds to whichever hand enters the field
of view, so in a shared indoor space it admits commands from bystanders.
Predictable interaction therefore requires two rarely combined capabilities:
identity-aware authorization, so that only an enrolled operator's commands are
honoured, and deterministic arbitration between teleoperation, tracking and
safety behaviours.

This paper presents \method{IGate}, an identity-gated gesture control stack for a drone with a monocular camera, and evaluates it in-flight on the DJI Tello EDU.
The design is conventional in its parts: MediaPipe landmarks, an RBF-SVM gesture
classifier, an ArcFace-trained face embedding, and a hierarchical state machine.
Our contribution is the in-flight evaluation, and a choice of components driven by cheap computation at acceptable precision.
Each component is characterised both by the offline metric conventionally
reported for it and by the rate it achieves through the drone's own video
pipeline, and for every component that measurement is taken in flight.

The size of the gap between the two is not uniform across the stack.
It is large where a component depends on image quality (video compression, motion blur, lighting): the identity gate loses
a factor of 23 in false rejection to the aircraft's imagery at a fixed
threshold.
It is small where a component depends on evaluation protocol: gesture accuracy
moves from 0.997 to 0.850 almost entirely through the choice of split.
Diagnosing which is at work requires per-frame logging dense enough to
reconstruct a trial's internal state after the fact, released here with the
system.

The contributions are as follows.

\begin{itemize}
\item \method{IGate}, an identity-gated human-drone interaction stack in which
      gesture and tracking commands are filtered by session-level face
      verification and arbitrated by a deterministic state machine, with a local language model excluded from actuation.
\item A three-protocol evaluation of the gesture classifier giving 0.997, 0.910
      and 0.783, and a hover-locked in-flight comparison of both deployed
      classifiers at matched standoff giving 0.850 against 0.651.
\item An operational characterisation of the identity gate: false rejection
      through the aircraft video pipeline and an authorization envelope in
      operator standoff, reported against the offline equal error rate of the
      same model and decomposed into the cost of the imagery and the cost of the
      operating point.
\item Six deployment behaviours that appear only in flight, three of them
      corrected and reported with paired measurements.
\end{itemize}

Every quantitative claim is rederived from the released logs by an accompanying
audit script, which reports any figure it cannot reproduce.

\section{Related Work}
\label{sec:related}

\begin{table}[t]
\caption{Positioning against representative gesture-controlled UAV systems.
``Operational'' means measured through the aircraft's own video pipeline.
$^{\dagger}$In-flight gesture accuracy and in-flight false rejection.}
\label{tab:related}
\centering
\footnotesize
\setlength{\tabcolsep}{4pt}
\renewcommand{\arraystretch}{1.02}
\begin{tabular}{@{}lcccc@{}}
\toprule
& Identity & Explicit & Cross-sess. & Operational \\
& gate & arbitration & eval. & rates$^{\dagger}$ \\
\midrule
Latif \emph{et al.}~\cite{latif2022}      & --  & --  & --  & -- \\
HGIC~\cite{hu2024hgic}                    & --  & yes & --  & -- \\
Taylor \emph{et al.}~\cite{taylor2025}    & --  & yes & --  & -- \\
Seidu \& Lawal~\cite{seidu2024}           & yes & --  & --  & -- \\
Varga~\cite{varga2026}                    & --  & --  & yes & -- \\
\midrule
This work                                 & yes & yes & yes & 0.850\,/\,19.3\% \\
\bottomrule
\end{tabular}
\end{table}

Gesture control for UAVs is well established.
Wachs \emph{et al.}~\cite{wachs2011} set out the requirements for hand-gesture
interfaces, and later systems map landmarks or learned features to flight
commands~\cite{zhang2020,latif2022,hu2024hgic,taylor2025,abdalla2025edge},
against dedicated datasets such as \mbox{UAV-GESTURE}~\cite{perera2018uavgesture}.
These systems report classification accuracy on curated datasets and
demonstrate flight qualitatively; they do not report what the recogniser
achieves through the aircraft's own video link, and they treat any detected
hand as a valid operator.

Face recognition supplies the missing authorization.
ArcFace~\cite{deng2019} is the standard margin-based objective and is reported
by equal error rate on still-image benchmarks such as LFW~\cite{huang2008}.
That characterisation is known to travel poorly: Cheng \emph{et
al.}~\cite{cheng2018survface} report a large drop between benchmark figures and
native surveillance imagery.
Whether it survives a rolling-shutter camera, an H.264 encoder and a 2.4~GHz
link is not usually asked, and Section~\ref{sec:identity} finds that it does
not.

On evaluation protocol, Varga~\cite{varga2026} shows that random splits over
frames from continuous recordings overstate hand-gesture accuracy, and argues
for subject-independent partitioning.
Section~\ref{sec:gesture} reaches a consistent conclusion from
session-independent partitioning, which removes a different leakage axis: all
four sessions here are one operator, so every figure in this paper is
within-subject and is an upper bound on cross-subject performance.
Language models have been proposed as robot planners~\cite{ahn2022saycan} and
constrained by guardrails~\cite{ravichandran2025}; \method{IGate} excludes the
model from actuation entirely.

\section{Method}
\label{sec:method}

\begin{figure*}[t]
\centering
\resizebox{\textwidth}{!}{

\begin{tikzpicture}[x=1mm,y=1mm]

\tikzset{
  blk/.style   = {draw, line width=0.4pt, inner sep=3pt, align=center,
                  minimum height=6.5mm, font=\scriptsize},
  wide/.style  = {blk, text width=20mm},
  narrow/.style= {blk, text width=17mm},
  gate/.style  = {wide, fill=black!4},
  aux/.style   = {wide, draw=black!35, text=black!55},
  flow/.style  = {-{Stealth[length=1.6mm,width=1.4mm]}, line width=0.4pt},
  aside/.style = {flow, densely dashed, draw=black!55},
  lbl/.style   = {font=\fontsize{6}{7}\selectfont, inner sep=1.5pt},
  grp/.style   = {draw=black!25, line width=0.3pt, rounded corners=1pt,
                  inner sep=0pt},
}

\node[blk, text width=25mm, minimum height=34mm] (air) at (0,0) {DJI Tello EDU\\[3pt]\fontsize{6}{7.5}\selectfont camera\\ IMU, barometer\\ motors};
\node[lbl, above=1.2mm of air] {\textsc{aircraft}};
\node[grp, fit={(38,-19) (177,17)}] (host) {};
\node[lbl, above=0.6mm of host.north] {\textsc{host}};
\node[wide] (buf) at (52,7.5) {latest-frame buffer\\\fontsize{6}{7}\selectfont thread, depth 1};
\node[narrow] (hand) at (79,7.5) {MediaPipe Hands\\\fontsize{6}{7}\selectfont every 2nd frame};
\node[narrow] (face) at (79,-7.5) {MediaPipe face\\\fontsize{6}{7}\selectfont every 4th frame};
\node[narrow] (clf) at (105,7.5) {gesture classifier\\\fontsize{6}{7}\selectfont rule \textbar{} RBF-SVM};
\node[narrow, fill=black!5] (gate) at (105,-7.5) {identity gate\\\fontsize{6}{7}\selectfont $\tau_{\mathrm{on}}{=}0.55$, $\Delta_{\max}{=}0.35$\,s};
\node[narrow, fill=black!5] (fsm) at (136,0) {arbitration\\\fontsize{6}{7}\selectfont priority order};
\node[narrow] (rc) at (163,0) {RC command\\\fontsize{6}{7}\selectfont 10\,Hz};
\node[aux, text width=22mm] (llm) at (136,12.5) {reason-only LLM\\\fontsize{6}{7}\selectfont local Qwen2.5-0.5B};
\draw[flow] (air.east |- buf) -- (buf.west);
\node[lbl, above=0.3mm] at (25.80,7.5) {video};
\node[lbl, below=0.3mm] at (25.80,7.5) {UDP 11111};
\draw[flow] (buf.east) -- (hand.west);
\draw[flow] (hand.east) -- (clf.west);
\draw[flow] (buf.south) -- (52,-7.5) -- (face.west);
\draw[flow] (face.east) -- (gate.west);
\draw[line width=0.4pt] (clf.east) -- (122,7.5);
\draw[line width=0.4pt] (gate.east) -- (122,-7.5);
\draw[line width=0.4pt] (122,7.5) -- (122,-7.5);
\draw[flow] (122,0) -- (fsm.west);
\fill (122,0) circle (0.45);
\draw[flow] (fsm.east) -- (rc.west);
\draw[aside] (fsm.north) -- (llm.south);
\draw[flow] (air.east |- 0,-15) -- (136,-15) -- (fsm.south);
\node[lbl, above=0.3mm] at (25.80,-15) {telemetry};
\node[lbl, below=0.3mm] at (25.80,-15) {UDP 8890};
\draw[flow] (rc.east) -- (183,0) -- (183,-25) -- node[lbl, pos=0.42, above=0.3mm] {command} node[lbl, pos=0.42, below=0.3mm] {UDP 8889} (0,-25) -- (air.south);
\end{tikzpicture}}
\caption{\method{IGate}. The aircraft carries no computation: perception,
authorization and arbitration run on the host, and every stage is logged per frame. The language model sits off this path entirely.}
\label{fig:arch}
\end{figure*}
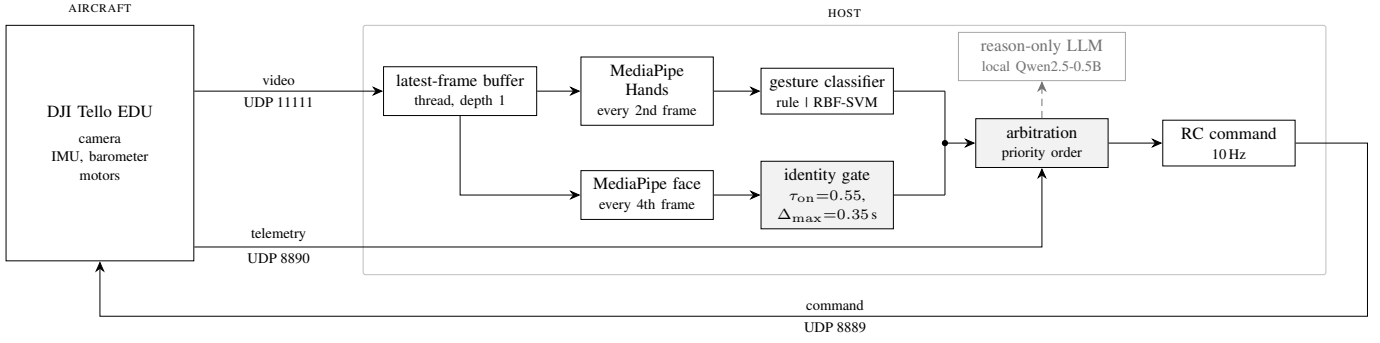

\begin{table}[t]
\caption{Parameters Section~\ref{sec:method} specifies and Section~\ref{sec:results} measures. Values are generated from the deployed configuration.}
\label{tab:design}
\centering
\footnotesize
\setlength{\tabcolsep}{4pt}
\renewcommand{\arraystretch}{1.05}
\begin{tabular}{@{}llc@{}}
\toprule
Parameter & Value & Measured in \\
\midrule
$\theta_{\mathrm{scale}}$ on $\Delta s_t$ & 0.18 & \S\,\ref{sec:behaviours} \\
Face-follow target $\rho^{\star}$ & 0.075 of frame & \S\,\ref{sec:identity} \\
$\tau_{\mathrm{on}}$ & 0.55 & \S\,\ref{sec:identity} \\
$\tau_{\mathrm{off}}$, $k$ & 0.45, 3 & \S\,\ref{sec:behaviours} \\
$\Delta_{\max}$ crop freshness & 0.35~s & \S\,\ref{sec:behaviours} \\
Mode hold, release & 1.2~s, 0.8~s & \S\,\ref{sec:results} \\
Search sweep & 28~s & \S\,\ref{sec:behaviours} \\
Battery failsafe & 15\% & \S\,\ref{sec:behaviours} \\
\bottomrule
\end{tabular}
\end{table}

\subsection{Platform and Perception}

The platform is a DJI Tello EDU (TLW004), an 87~g quadrotor with a stabilised
RGB camera and no obstacle-avoidance sensors; perception, telemetry and
arbitration run on a host PC (Intel Core i7-9750H, CPU-only inference) over the
three UDP channels of Fig.~\ref{fig:arch}.
The host decodes video in a separate thread into a single-slot buffer, which
bounds staleness at the cost of throughput, and the age of each frame at
consumption is logged, since a host-side latency measurement that omits it
describes the wrong system.
Both detectors are throttled, hands every second frame and faces every fourth.

\begin{figure}[t]
\centering
\begin{minipage}[b]{0.51\columnwidth}
\centering
\includegraphics[width=\linewidth]{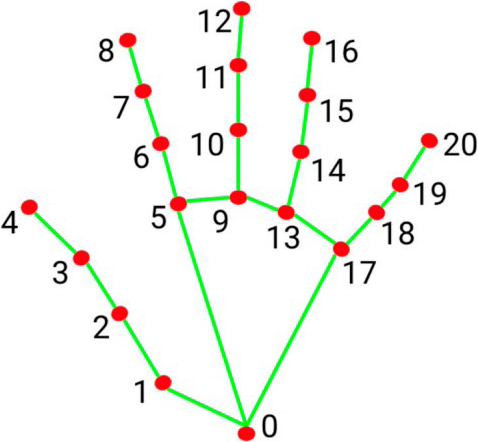}
\end{minipage}\hfill
\begin{minipage}[b]{0.44\columnwidth}
\centering
\includegraphics[width=\linewidth]{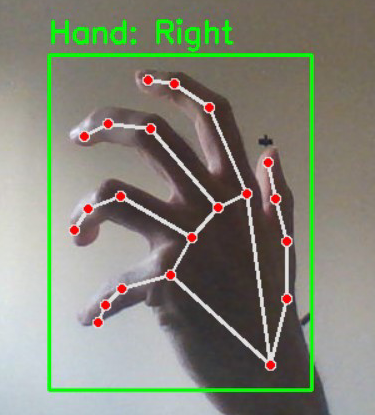}
\end{minipage}
\caption{The 21-point hand skeleton \method{IGate} classifies (left) and the
runtime overlay the operator sees (right). The numbered joints are the indices
used throughout: 0 the wrist, 4 the thumb tip, 5 and 8 the index
metacarpophalangeal joint and tip that carry the rule's directional vector, and
20 the little-finger tip that separates \texttt{LEFT} from \texttt{BACK}
(Section~\ref{sec:behaviours}).}
\label{fig:skeleton}
\end{figure}

MediaPipe Hands produces 21 landmarks per detected hand, flattened to
\begin{equation}
\mathbf{f}_t=[x_1,y_1,z_1,\dots,x_{21},y_{21},z_{21}]^{\!\top}\in\mathbb{R}^{63}.
\label{eq:feat}
\end{equation}
Coordinates are normalised to the image frame, so the representation is
invariant to where the hand sits in view but \emph{not} to how large it appears;
Section~\ref{sec:gesture} finds that residual scale dependence to be the
strongest single predictor of cross-session transfer.
The geometric rule smooths $\mathbf{f}_t$ with an exponential moving average at
$\alpha=0.35$ and resets it when the hand is lost; the learned classifier
receives $\mathbf{f}_t$ unsmoothed.

\subsection{Two Gesture Vocabularies}

\method{IGate} carries two gesture classifiers, and they decode \emph{different
vocabularies}: the rule reads where the index finger points, the classifier
reads the configuration of the whole hand.
A pose meaning \texttt{FORWARD} to one is not a degraded \texttt{FORWARD} to the other but an
input it has no reading for, so every comparison in Section~\ref{sec:results}
cues each classifier on the vocabulary it decodes.

With $\mathbf{p}_{\mathrm{tip}}$ and $\mathbf{p}_{\mathrm{mcp}}$ the index tip
and metacarpophalangeal joint, the rule's directional vector is
$\mathbf{v}=\mathbf{p}_{\mathrm{tip}}-\mathbf{p}_{\mathrm{mcp}}$, and a lateral
or vertical component is emitted when the corresponding element of $\mathbf{v}$
exceeds $\theta_{\mathrm{dir}}=0.10$ in normalised units.
Depth is inferred not from a pose but from a \emph{change}: with $A_t$ the
landmark bounding-box area,
\begin{equation}
\Delta s_t=\frac{A_t-A_{t-1}}{A_{t-1}},
\label{eq:deltas}
\end{equation}
and \texttt{FORWARD} or \texttt{BACK} is emitted when $|\Delta s_t|$ exceeds
$\theta_{\mathrm{scale}}=0.18$.
Equation~\eqref{eq:deltas} is not normalised by $\Delta t$, so the same physical
motion yields different values at different frame rates.
The learned classifier standardises $\mathbf{f}_t$ and applies a support vector
machine with a radial basis function kernel, $C=10$ and $\gamma$ set by the
scale heuristic, over seven classes, fitted on a stratified 75/25 split at
seed~42.

The classifier is selected explicitly at launch, validated before the aircraft
is armed, and recorded in the run manifest with the model's SHA-256; a released
regression test fails on any run whose manifest and per-frame latency columns
disagree.

\subsection{Identity Gate and Arbitration}

At startup the operator enrols: $N=20$ accepted face crops (bounding boxes) are embedded and averaged
into a template $E_{\mathrm{auth}}$, and each later crop $c_t$ is compared with it by cosine similarity,
$S_t=\langle \phi(c_t),E_{\mathrm{auth}}\rangle$, both vectors being unit length.
The embedding network is frozen; enrolment only averages its outputs.
Authorization $a_t$ is a Schmitt trigger rather than a per-frame comparison,
\begin{equation}
a_t=
\begin{cases}
1, & S_t\ge\tau_{\mathrm{on}},\\
0, & S_t<\tau_{\mathrm{off}} \text{ for } k \text{ consecutive frames},\\
a_{t-1}, & \text{otherwise},
\end{cases}
\label{eq:gate}
\end{equation}
with $\tau_{\mathrm{on}}=0.55$, $\tau_{\mathrm{off}}=0.45$ and $k=3$.
The opening threshold sits well above the 0.18 that balances the two error rates
offline, because a false accept hands control to the wrong person while a false
reject only costs a repeated gesture (Section~\ref{sec:identity}).
MediaPipe detects all faces in the frame and supplies their bounding boxes, whose padded crop is resized to
$112\times112$ and embedded by $\phi(\cdot)$, a
MobileFaceNet~\cite{chen2018mobilefacenets} trained on WebFace600K under the
ArcFace additive angular margin loss~\cite{deng2019}.
The embedding is therefore used outside the pipeline it was trained in, which
aligns each crop onto five facial landmarks and pairs the network with a
RetinaFace~\cite{deng2020retinaface} detector; Section~\ref{sec:identity} treats that as one untested
candidate for the operational false rejection rate.

Detection and recognition carry different time constants and the two are not
interchangeable.
Arbitration holds its face predicate across detection gaps for 2.0~s so that
mode selection does not flicker on a low-frame-rate stream; identity accepts a
crop only if its detection is younger than $\Delta_{\max}=0.35$~s, because
re-embedding a stale box feeds the network whatever now occupies a region the
face has left.

A gesture is admitted when a hand is present and $a_t=1$.
These are two independent tests over the frame, not an association: the
perception path returns a single hand and never binds it to the authorized face,
so the gate establishes operator presence rather than command provenance.

Face following uses the detection box, not the embedding: the aircraft centres
the box and holds its area at a fixed fraction of the frame, which
Section~\ref{sec:identity} calibrates to a 0.94~m standoff. The embedding
decides whether to follow, the box where to fly.

A state machine selects one mode per control tick by the first satisfied
condition in the fixed order: battery failsafe, gesture, face following, bounded search, hover.
Battery at or below 15\% forces an irreversible landing.
A mode is held for at least 1.2~s and released only after its signal has been
absent for 0.8~s.
With nothing seen for 10~s the aircraft begins a yaw sweep of 28~s, one full
rotation at the deployed rate of approximately 13\textdegree/s.

Every command is computed by \eqref{eq:gate} and the state machine.
A structured summary of each decision then goes to a local Qwen2.5-0.5B
instance, which returns one sentence for the log but cannot reach the actuation path.

\section{Experimental Setup}
\label{sec:setup}

Each launch writes a timestamped run directory containing telemetry, decision,
per-frame performance and trial logs, with a manifest recording configuration,
\textit{Git} commit, library versions and host.
Two conventions make the latency figures meaningful: a throttled stage
contributes a sample only on frames where it executed, and frames are
deduplicated by sequence number before any rate is computed.

Four sessions of seven gesture classes (\texttt{CENTER}, \texttt{LEFT}, \texttt{RIGHT}, \texttt{UP}, \texttt{DOWN}, \texttt{FORWARD},
\texttt{BACK}) were captured at 400 images per class under deliberately different
conditions rather than as replicates.
Session A is the reference condition, daylight against a uniform blue wall
separating hand from background in intensity and hue.
B and C were recorded 
in two domestic rooms under artificial
light against cluttered backdrops, and D in daylight against a cluttered room,
the cell that separates the two factors.
After landmark extraction the usable samples are 2715, 2345, 2613 and 2784.

A flight trial counts only if the aircraft was airborne within the trial window,
verified from the command log; the rule is uniform and does not select on
outcome. Of 150 flight trials one was excluded, having been opened after a
failsafe landing and before the next takeoff, leaving 149 flown among 270 valid trials.
All 149 produced the specified mode transition, as did 61 of 61 webcam-harness
trials in the deployed configuration.
With no trial failing, this result is at ceiling: it shows the arbitration
layer behaves as designed, but cannot tell correct perception beneath it from
incorrect.

Latency is measured over 557~s airborne and 12{,}419 frames.
Median end-to-end camera-to-command latency is 18.6~ms (95th percentile
55.2~ms) at 22.3~fps, of which frame age at consumption contributes a median
4.0~ms and a 95th percentile of
30.5~ms.
Stage costs are higher in flight than on the ground, hand detection rising from
24.4 to 26.7~ms and face embedding from 9.0 to 13.0~ms.
Gesture inference costs a mean 0.84~ms for the SVM and 0.07~ms for the rule, so
the choice between them is not a latency trade-off.

\section{Results}
\label{sec:results}

\subsection{Gesture Classification at Three Protocols}
\label{sec:gesture}

The classifier is scored three ways, each stricter than the last: a
\emph{within-session} split; a \emph{held-out session} split, trained on three
sessions and tested on the fourth; and an \emph{operational} measurement on
frames the aircraft itself encoded. The metric is per-frame accuracy over the
seven classes.

Evaluated within each session under a stratified random split, the SVM reaches an accuracy of 
0.997, 0.988, 1.000 and 1.000 on sessions A, B, C and D.
The protocol does not separate the reference condition from the three degraded
ones: it returns 1.000 on both C and D and its lowest figure on A, the condition
in which hand and background are most separable.
A protocol that ranks the hardest capture conditions above the easiest is
characterising the recording rather than the classifier.
The mechanism is visible in the sampling density: within session A accuracy
rises monotonically with the frames retained per class, 0.9486 at 100, 0.9840 at
250, 0.9971 at 400.
Frames captured milliseconds apart during a continuous recording of a static
pose are near-identical in landmark space, so a random split distributes
near-duplicates across both partitions.
That is the signature of leakage, and is consistent with the
subject-independence analysis of Varga~\cite{varga2026}.

Leave-one-session-out (Group K-fold) cross-validation gave 0.793, 0.882, 0.978 and 0.988, a mean of 0.910.
Transfer is markedly asymmetric: A reaches 0.815, 0.842 and 0.930 on B, C and D as a training set, whereas C reaches only 0.471 and 0.509 on A and B.
A clean backdrop is therefore the more valuable training condition and the more
misleading test condition.
Grouping the twelve ordered pairs by what changes between their two sessions
gives 0.739 where neither factor moves (2 pairs), 0.784 for background alone
(2), 0.880 for illumination alone (4) and 0.709 where both move (4).
The two single-factor groups order as expected but their ranges overlap almost
completely, 0.638--0.930 against 0.737--0.977, and the difference is not
significant ($p=0.53$); at two against four the smallest attainable $p$ is
0.067, so the design cannot reach significance even in principle.
A post-hoc analysis suggests the axis it misses: the fraction of a test
session's hand-scale distribution lying inside the training session's
5th--95th percentile range correlates with transfer at Pearson $r=0.76$ over the 12
pairs, against 0.30 for shared background and $-0.09$ for shared illumination.
This is exploratory, chosen after seeing the matrix at $n=12$, and is a caveat
on the grouping rather than a replacement for it.

A third measurement scores the classifier on frames the aircraft encoded,
transmitted and the host decoded.
With the drone powered and streaming on the ground and its propellers removed,
a cued protocol named one of the seven classes, allowed 5~s to form it and
recorded 6~s, over 42 holds in six rounds of randomised class order.
The SVM reaches 0.783 over all 2947 cued frames and 0.809 over the 42 holds, the
hold being the independent unit because the operator was cued once and formed
the gesture once.
A hand was detected in every frame, so the loss is entirely misclassification.

The three protocols therefore give 0.997, 0.910 and 0.783, each stricter than
the last.

\subsection{Both Classifiers in Flight}
\label{sec:inflight}

\begin{table}[t]
\caption{Gesture classification in flight $\uparrow$, hover-locked cued capture. Recall is unconditioned; the bracketed figure conditions on detection. Per-hold scores each of the 42 holds by majority vote.}
\label{tab:inflight}
\centering
\footnotesize
\setlength{\tabcolsep}{3pt}
\renewcommand{\arraystretch}{1.02}
\begin{tabular}{@{}lrrrrrr@{}}
\toprule
& \multicolumn{3}{c}{RBF-SVM} & \multicolumn{3}{c}{Rule} \\
\cmidrule(lr){2-4}\cmidrule(lr){5-7}
Class & Recall & Prec. & $F_1$ & Recall & Prec. & $F_1$ \\
\midrule
CENTER & 0.988 & 0.941 & 0.964 & 0.881 & 0.567 & 0.690 \\
LEFT & 0.980 & 0.736 & 0.841 & 0.999 & 1.000 & 0.999 \\
RIGHT & 0.985 & 0.685 & 0.808 & 0.968 & 0.997 & 0.982 \\
UP & 0.999 & 1.000 & 0.999 & 0.854 & 0.548 & 0.668 \\
DOWN & 0.920 & 1.000 & 0.958 & 0.908 & 1.000 & 0.952 \\
FORWARD & 0.777 (0.858) & 1.000 & 0.874 & 0.000 & 0.000 & 0.000 \\
BACK & 0.336 & 0.859 & 0.483 & 0.008 & 0.750 & 0.017 \\
\midrule
Per frame & \multicolumn{3}{c}{0.850 [0.840, 0.860]} & \multicolumn{3}{c}{0.651 [0.638, 0.664]} \\
Per hold & \multicolumn{3}{c}{0.905 [0.779, 0.962]} & \multicolumn{3}{c}{0.714 [0.564, 0.828]} \\
Detection & \multicolumn{3}{c}{0.986} & \multicolumn{3}{c}{0.984} \\
\bottomrule
\end{tabular}
\end{table}

A final protocol puts both classifiers in the air under the identical cue.
Cued holds are not flyable in a domestic room with actuation live, since a
sustained directional command translates roughly 3.9~m in six seconds, so the
captures use a hover lock: perception, gating, arbitration and the command
computation run and are logged exactly as deployed, while the transmitted
radio-control command is held at zero.
Suppression is applied at the transmit boundary, so nothing upstream branches on
it and the logged command is the one the aircraft would have received.

Each classifier was cued on the vocabulary it decodes over 42 holds in six rounds of
randomised class order, at a matched standoff inside the authorization envelope
of Section~\ref{sec:identity}: median 0.85~m for the SVM and 0.89 and
0.81~m for the two rule captures.
These 84 holds are a separate protocol and are not among the 270 trials.
Table~\ref{tab:inflight} reports the result.
The SVM reaches 0.850 per frame [0.840, 0.860] and 0.905 per hold, the rule
0.651 and 0.714, with hand detection above 0.98 in both.
One asymmetry bounds the comparison: the moving average is applied inside the
rule rather than in the shared perception path, so the rule is scored on
smoothed landmarks and the SVM on raw ones.
The smoothing favours the rule, so it cannot account for the direction of the
gap.

The gesture classifier does not degrade in flight as the identity gate does.
It reaches 0.783 on the ground and 0.850 airborne through the aircraft's optics,
against 0.910 held out across sessions and 0.997 within session: what governs
the gesture figure is the evaluation protocol, not the platform.

The comparison was specified before it was flown.
A preregistration fixed the primary metric as per-hold accuracy on the depth
channel, the sample size at 42 holds per classifier with the all-class comparison
declared underpowered and descriptive, the frozen model and its SHA-256, and an
exclusion rule admitting only declared operator error.
It recorded one directional prediction, that the SVM's \texttt{RIGHT} would collapse into
\texttt{BACK} as it had on the ground, at $F_1$ 0.433.
\emph{That prediction was disconfirmed.}
In flight the SVM reads \texttt{RIGHT} correctly on 0.985 of cued frames and \texttt{BACK} never;
the confusion reversed direction, \texttt{BACK} becoming the error source rather than its
target and dividing between \texttt{LEFT} at 0.33 and \texttt{RIGHT} at 0.30.
The preregistration carries two caveats: it was written to disk before the
first flight but never placed under version control, and its cued protocol
assumes an actuation revised to the hover lock above.

No static class falls below 0.92 recall on the SVM, \texttt{DOWN} reaching that figure
exactly; on $F_1$ two fall short, \texttt{LEFT} at 0.841 and \texttt{RIGHT} at 0.808, because \texttt{BACK}
leaks into both and recall does not show it.
Two of five reach 0.92 on the rule, \texttt{CENTER}, \texttt{UP} and \texttt{DOWN} falling short at 0.881,
0.854 and 0.908.
Of the 0.199 gap in per-frame accuracy, 0.162 --- 82\% --- is contributed by
\texttt{FORWARD} and \texttt{BACK} alone; the remaining 0.037 falls on \texttt{UP} and \texttt{CENTER}.

\subsection{Identity Verification}
\label{sec:identity}

\begin{table}[t]
\caption{False rejection of the enrolled operator $\downarrow$, in per cent, at the offline equal-error threshold and at the deployed one.}
\label{tab:frr}
\centering
\footnotesize
\setlength{\tabcolsep}{4pt}
\renewcommand{\arraystretch}{1.02}
\begin{tabular}{@{}lrr@{}}
\toprule
Condition & Frames & FRR (\%) \\
\midrule
Offline still images, $\tau=0.18$ & 13{,}410 & 0.32 \\
Drone in flight, $\tau=0.18$ & 684 & 7.3 \\
Drone in flight, $\tau=0.55$, per frame & 684 & 35.1 \\
Drone in flight, $\tau=0.55$, deployed gate & 684 & 19.3 \\
Drone on ground, $\tau=0.55$, per frame & 2792 & 16.7 \\
Drone on ground, $\tau=0.55$, deployed gate & 2792 & 9.2 \\
\bottomrule
\end{tabular}
\end{table}

\begin{figure}[t]
\centering
\includegraphics[width=0.92\columnwidth]{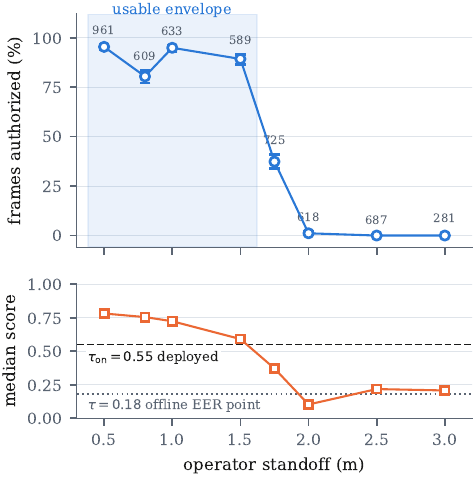}
\caption{Authorization rate $\uparrow$ against operator standoff, frame counts
annotated, Wilson intervals. Below: median similarity against the two
thresholds.}
\label{fig:envelope}
\end{figure}

\begin{figure*}[t]
\centering
\includegraphics[width=0.94\textwidth]{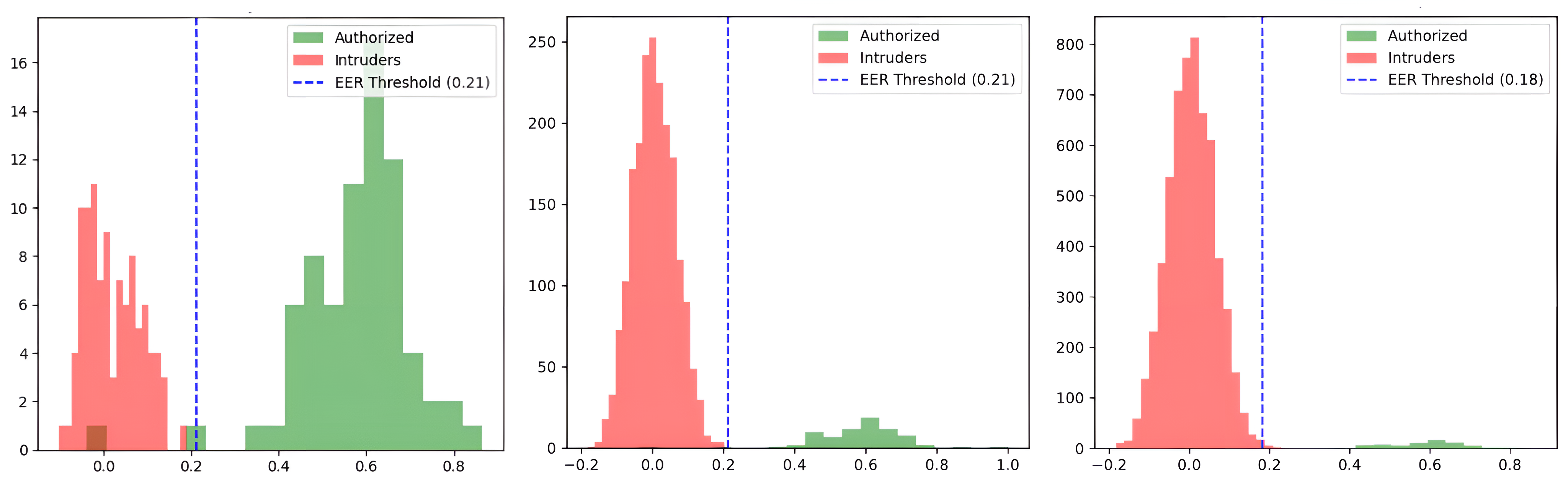}
\caption{Similarity scores of the enrolled operator against impostors, as the
negative set grows: 100 webcam images of other individuals, then $N=4{,}678$,
then the full $N=13{,}410$ including Labeled Faces in the
Wild~\cite{huang2008}. The dashed line is the equal-error threshold, which holds
at 0.21 until the negative set is enlarged and then settles at 0.18 at an equal
error rate of 0.32\%. The two populations stay separated throughout; what
Section~\ref{sec:identity} measures is what happens to that separation when the
imagery comes from the aircraft.}
\label{fig:offlineeer}
\end{figure*}

The embedding is characterised first in the manner conventional for the
component.
A positive set of 76 images of the enrolled operator was scored against a
negative set combining 100 webcam images of other individuals and
approximately 13{,}000 images from Labeled Faces in the Wild~\cite{huang2008}.
Against the full combined set of 13{,}410 the equal error rate is 0.32\%, at a
threshold of $\tau=0.18$.

That benchmark is not the operating point the aircraft runs at; it was performed for validation.
The deployed gate opens at $\tau_{\mathrm{on}}=0.55$, chosen for impostor margin
rather than for equal-error balance, so comparing the two figures directly
confounds a change of imagery with a change of threshold.
Table~\ref{tab:frr} separates them by holding the threshold fixed.
At the offline threshold, moving from curated stills to the aircraft's own video
raises false rejection from 0.32\% to 7.3\%, a factor of 23.
H.264 compression, motion blur, variable illumination and a moving camera are
the obvious candidates; a fourth is that the embedding is used outside the
pipeline it was trained in, on an unaligned crop from a different detector
(Section~\ref{sec:method}).
Which dominates is not resolved here, since separating them needs a capture that
varies one at a time.

Raising the threshold to the deployed 0.55 raises false rejection again to
35.1\%.
The choice of operating point therefore costs more than the change of imagery
does.
The Schmitt trigger of \eqref{eq:gate} recovers part of that, returning 19.3\%
against 35.1\% for the same frames thresholded independently, so hysteresis is
worth 16 points of false rejection measured on identical data.
On the ground, inside the envelope below, false rejection is 16.7\% per frame
and 9.2\% through the deployed gate over 2792 frames, both below the
corresponding in-flight rates, which is the expected ordering.

Fig.~\ref{fig:envelope} reports authorization against declared standoff over a
cued ground sweep of 5{,}103 scored frames at eight marked distances.
Between 0.5 and 1.5~m the gate admits 80--95\% of frames; at 1.75~m it admits
37\%, and at 2.0~m and beyond essentially none.
The usable envelope is therefore 0.5--1.5~m.
The same sweep calibrates the face-following standoff against the gate's: over
the 4{,}861 frames at marks inside the envelope the median of $\sqrt{\rho_t}\,d$ is
0.258~m, with $\rho_t$ the face box as a fraction of frame area and $d$ the
declared distance, so the follower's target $\rho^{\star}=0.075$ sits at 0.94~m,
near the middle of the range the gate admits.
The offline benchmark cannot supply this, since it relates a control parameter
to a physical standoff through the deployed optics.

Beyond the envelope the failure changes character.
Median apparent face size falls with distance as far as 1.75~m and then rises
again, from 132~px to 165, 211 and 210~px at 2.0, 2.5 and 3.0~m, which no
receding face can do.
Beyond 2.0~m the scores have mean 0.177 and standard deviation 0.082,
statistically the background population identified at $0.193\pm0.033$: the
detector is returning false positives on the backdrop and the gate is embedding
them.
Facial identity verification refuses them correctly, but arbitration is still told a face is
present.

\subsection{Deployment Behaviours}
\label{sec:behaviours}

Six behaviours appeared only in flight.
Three were corrected and are reported with paired measurements on the same
protocol.

\emph{Stale bounding boxes.}
The arbitration layer held its face predicate across detection gaps, and the
identity path re-embedded the same held box.
Restricting identity crops to detections under $\Delta_{\max}=0.35$~s old drops
the share of frames scoring in the background population from 56\% to 20\% and
raises the authorized share from 41\% to 68\%, while spurious search decisions
fall from 130 to 4.
Two correct components had shared one temporal predicate.

\emph{Authorization hysteresis.}
The Schmitt trigger of \eqref{eq:gate} reduces the mode-switch rate from 78.5 to
15.7 per minute against a bare per-frame threshold on identical data.
It is not free: under the bare threshold no impostor was ever authorized, and
with hysteresis 5.6\% of impostor frames were, across 20 trials in which the
impostor was otherwise rejected.

\emph{Depth commands.}
Expressed as a static pose and read by the SVM, \texttt{FORWARD} and \texttt{BACK} reach 0.777 and
0.336 in flight against 0.000 and 0.008 for the motion cue, on the same aircraft
at the same standoff under the same protocol.
The pose formulation removes the motion cue's pathology outright.
A motion cue has no steady state: the correct component fired in every one of
the twelve depth holds but occupied 9\% of the window, the return stroke
supplies the opposite command unless made slower than the threshold, and 67\% of
\texttt{BACK} holds emitted a spurious \texttt{FORWARD}.
The $\Delta t$ defect of \eqref{eq:deltas} is therefore a consequence of
choosing a motion cue rather than an independent fault.

\emph{Vocabulary packing.}
\texttt{BACK} does not follow \texttt{FORWARD}, and its 0.336 is a second and independent problem.
Three of the seven trained poses share a closed fist and are distinguished by a
single extended digit: \texttt{LEFT} by the little finger, \texttt{RIGHT} by the thumb, \texttt{BACK} by
neither.
Over the in-flight frames cued \texttt{BACK}, little finger extension normalised by palm
length is 0.884 on the frames read as \texttt{LEFT} against 0.551 on those read
correctly; genuine \texttt{LEFT} poses measure 0.821, so on the misread holds the
estimator reported a little finger \emph{more} extended than on the class it
belongs to.
The failure is one of landmark estimation, not of classification, and it follows
from a vocabulary whose classes are separated by less than the estimator
resolves.

\emph{Search coverage and battery failsafe.}
Measured from yaw telemetry, the deployed sweep covers a median 318\textdegree{}
over 22 episodes.
Its 28~s duration is set by the yaw rate: at the configured
$\approx$13\textdegree/s a full rotation requires 28~s, and an earlier 5~s
specification covered a median 60\textdegree{} for that reason.
Stepping the battery threshold down after each activation produced eleven
consecutive automatic landings between 91\% and 21\% charge in one 580~s flight.

\emph{Radio environment.}
Stream rate varied between 4 and 24~fps across 31 runs, and battery charge does
not explain it (pooled Pearson $r=-0.08$); venue does.
In a university building with dense access-point coverage a ground test achieved
21.8~fps with no reconnections, whereas flight in the same room on the same link
gave 8.4~fps with 48 reconnections and SDK command timeouts while airborne,
including a failed emergency stop.
Motor interference, higher current draw and a rotating antenna occur only in
flight, so bench evaluation does not expose this constraint.

\emph{Reason-only language model.}
In one instrumented run the model was called 199 times, populating 490 decision
records.
Eleven calls returned a timeout, 5.5\% of calls and 7.1\% of records, and
latency averaged 1766~ms against a command interval of 100~ms.
Across all runs the verbatim low-battery sentence was emitted at charges as high
as 94\%, a state the telemetry contradicts.
Each figure measures one of the properties for which the model was excluded
from actuation.

\section{Discussion}

Three of the behaviours above share a structure: each arose between components
that were individually correct and individually characterised. The stale crop
came from two components sharing one temporal predicate; the authorization
chatter from a correct threshold feeding a correct state machine; the spurious
depth commands from a correct rule evaluated at an uncontrolled rate.
Vocabulary packing is not of that kind. There the estimator and the classifier
both behave as specified, and the fault lies in the specification itself.
What exposed them was not component-level evaluation, nor the scenario
validation in which every trial passed, but per-frame logging, latency
conditioned on the frames where work occurred, and operational rates recorded
alongside offline metrics.

The gap between a reported figure and a deployed one has a parallel in the
simulation-to-reality literature~\cite{aljalbout2025realitygap}; here both
measurements are taken on the same hardware, and the mismatch is between two
evaluation protocols over one component.
Only the absolute rates belong to this aircraft.
A 19.3\% false rejection rate, a 4--24~fps stream and a 0.5--1.5~m envelope
follow from one rolling-shutter camera, one H.264 encoder and one 2.4~GHz link.
The couplings, the specification error and the measurement practice do not move
with the hardware, and it is the last of these that carries furthest.

\emph{Limitations.}
The in-flight comparison suppresses actuation, so it does not measure the closed
loop in which a command moves the aircraft and changes the operator's framing.
The two classifiers are cued on different vocabularies, necessarily, so the
comparison is between deployed capabilities rather than classifiers on matched
input, and the operator knows which one is running.
The results characterise one enrolled operator across four sessions; extending
them across operators is the natural next study.
The gate establishes that an authorized operator is present, not that the
commanding hand is theirs, and binding them through person detection is the most
valuable extension.
The radio-environment result rests on two sites.

\section{Conclusion}

This paper proposed \method{IGate}, an identity-gated gesture control stack for
a low-cost indoor drone, and evaluated it through 270 logged trials of which 149
were flown.
Every valid flight trial produced the specified behaviour.

The offline and operational figures disagree, and by how much depends on what
the component rests on.
Gesture accuracy falls 0.997, 0.910, 0.783 across three successively stricter
protocols, almost entirely through the choice of split; face verification
operates at 19.3\% false rejection in flight against a 0.32\% offline equal
error rate, a gap that divides into the cost of the imagery and the cost of an
operating threshold three times the equal-error point.
Flown under a hover-locked cue at matched standoff the RBF-SVM reaches 0.850 and
the geometric rule 0.651, with 82\% of that gap on the depth channel.
Six deployment behaviours were quantified from the logs and three corrected with
paired measurements.
These results support reporting operational rates and conditioned per-stage
latency alongside offline benchmarks for systems of this class.

\section*{Ethics, Data and Code Availability}

Face images of the operator were collected by the operator; public face datasets
were used only for offline evaluation.
The enrolled embedding is held in volatile memory for the session and is not
persisted.
The system implements no liveness detection or anti-spoofing and is not a
biometric security mechanism: its threat model is prevention of accidental
control by bystanders after enrolment.
Indoor flight used propeller guards, conservative command limits and manual
supervision.
Implementation, instrumentation, all run logs and the table- and
figure-generation scripts will be released at
\url{https://github.com/Diyari-Fariq-M-salih/gesture-controlled-tello-drone}.
The authors thank the M2 Smart Aerospace and Autonomous Systems program at
\textit{University Paris-Saclay}.

\balance
\bibliographystyle{IEEEtran}
\bibliography{references}

@article{wachs2011,
  author  = {Wachs, Juan Pablo and K{\"o}lsch, Mathias and Stern, Helman and Edan, Yael},
  title   = {Vision-based hand-gesture applications},
  journal = {Communications of the ACM},
  volume  = {54}, number = {2}, pages = {60--71}, year = {2011},
  doi     = {10.1145/1897816.1897838}
}

@inproceedings{zhang2020,
  author    = {Zhang, Fan and Bazarevsky, Valentin and Vakunov, Andrey and Tkachenka, Andrei
               and Sung, George and Chang, Chuo-Ling and Grundmann, Matthias},
  title     = {{MediaPipe} Hands: On-device Real-time Hand Tracking},
  booktitle = {CVPR Workshop on Computer Vision for Augmented and Virtual Reality},
  year      = {2020}
}

@article{taylor2025,
  author  = {Taylor, B. and Allen, M. and Henson, P. and Gao, X. and Malik, H. and Zhu, P.},
  title   = {Enhancing Drone Navigation and Control: Gesture-Based Piloting,
             Obstacle Avoidance, and {3D} Trajectory Mapping},
  journal = {Applied Sciences}, volume = {15}, number = {13},
  pages   = {7340}, year = {2025},
  doi     = {10.3390/app15137340}
}

@inproceedings{deng2019,
  author    = {Deng, Jiankang and Guo, Jia and Xue, Niannan and Zafeiriou, Stefanos},
  title     = {{ArcFace}: Additive Angular Margin Loss for Deep Face Recognition},
  booktitle = {Proc. IEEE/CVF Conf. on Computer Vision and Pattern Recognition},
  pages     = {4690--4699}, year = {2019}, doi = {10.1109/CVPR.2019.00482}
}

@inproceedings{huang2008,
  author    = {Huang, Gary B. and Mattar, Marwan and Berg, Tamara and Learned-Miller, Erik},
  title     = {Labeled Faces in the Wild: A Database for Studying Face Recognition
               in Unconstrained Environments},
  booktitle = {Workshop on Faces in Real-Life Images}, year = {2008}
}

@article{varga2026,
  author  = {Varga, Domonkos},
  title   = {On the Evaluation Protocol of ``Gesture Recognition for {UAV}-based
             Rescue Operation based on Deep Learning'': A Subject-Independence
             Perspective},
  journal = {arXiv preprint arXiv:2602.17854},
  year    = {2026}
}

@article{hu2024hgic,
  author  = {Hu, Mengsha and Li, Jinzhou and Jin, Runxiang and Shi, Chao and
             Xu, Lei and Liu, Rui},
  title   = {{HGIC}: A Hand Gesture Based Interactive Control System for
             Efficient and Scalable Multi-{UAV} Operations},
  journal = {arXiv preprint arXiv:2403.05478},
  year    = {2024}
}

@incollection{latif2022,
  author    = {Latif, Bilawal and Buckley, Neil and Secco, Emanuele Lindo},
  title     = {Hand Gesture and Human-Drone Interaction},
  booktitle = {Intelligent Systems and Applications},
  series    = {Lecture Notes in Networks and Systems},
  volume    = {544},
  pages     = {299--308},
  year      = {2022},
  publisher = {Springer},
  doi       = {10.1007/978-3-031-16075-2_20}
}

@article{ravichandran2025,
  author  = {Ravichandran, Zachary and Robey, Alexander and Kumar, Vijay and
             Pappas, George J. and Hassani, Hamed},
  title   = {Safety Guardrails for {LLM}-Enabled Robots},
  journal = {IEEE Robot. Autom. Lett.},
  year    = {2026},
  note    = {arXiv:2503.07885}
}

@article{ahn2022saycan,
  author  = {Ahn, Michael and Brohan, Anthony and Brown, Noah and others},
  title   = {Do As {I} Can, Not As {I} Say: Grounding Language in Robotic Affordances},
  journal = {arXiv preprint arXiv:2204.01691},
  year    = {2022}
}

@article{seidu2024,
  author  = {Seidu, Idris and Lawal, Jafaar Olasunkanmi},
  title   = {Personalized Drone Interaction: Adaptive Hand Gesture Control with
             Facial Authentication},
  journal = {International Journal of Scientific Research in Science,
             Engineering and Technology},
  volume  = {11},
  number  = {4},
  pages   = {43--60},
  year    = {2024},
  doi     = {10.32628/IJSRSET241146}
}

@inproceedings{perera2018uavgesture,
  author    = {Perera, Asanka G. and Law, Yee Wei and Chahl, Javaan},
  title     = {{UAV-GESTURE}: A Dataset for {UAV} Control and Gesture Recognition},
  booktitle = {Proc. Eur. Conf. Comput. Vis. (ECCV) Workshops},
  year      = {2018}
}

@article{abdalla2025edge,
  author  = {Abdalla, Sousannah and Baidya, Sabur},
  title   = {{UAV} Control with Vision-based Hand Gesture Recognition over Edge-Computing},
  journal = {arXiv:2505.17303},
  year    = {2025}
}

@inproceedings{chen2018mobilefacenets,
  author    = {Chen, Sheng and Liu, Yang and Gao, Xiang and Han, Zhen},
  title     = {{MobileFaceNets}: Efficient {CNNs} for Accurate Real-Time Face
               Verification on Mobile Devices},
  booktitle = {Chinese Conf. Biometric Recognition (CCBR)},
  year      = {2018}
}

@inproceedings{deng2020retinaface,
  author    = {Deng, Jiankang and Guo, Jia and Ververas, Evangelos and
               Kotsia, Irene and Zafeiriou, Stefanos},
  title     = {{RetinaFace}: Single-Shot Multi-Level Face Localisation in the Wild},
  booktitle = {Proc. IEEE/CVF Conf. Comput. Vis. Pattern Recognit. (CVPR)},
  year      = {2020},
  pages     = {5203--5212}
}

@article{cheng2018survface,
  author  = {Cheng, Zhiyi and Zhu, Xiatian and Gong, Shaogang},
  title   = {Surveillance Face Recognition Challenge},
  journal = {arXiv:1804.09691},
  year    = {2018}
}

@article{aljalbout2025realitygap,
  author  = {Aljalbout, Elie and Xing, Jiaxu and Romero, Angel and
             Akinola, Iretiayo and Garrett, Caelan Reed and Heiden, Eric and
             Gupta, Abhishek and Hermans, Tucker and Narang, Yashraj and
             Fox, Dieter and Scaramuzza, Davide and Ramos, Fabio},
  title   = {The Reality Gap in Robotics: Challenges, Solutions, and Best Practices},
  journal = {Annu. Rev. Control Robot. Auton. Syst.},
  volume  = {9},
  year    = {2026},
  note    = {To appear; arXiv:2510.20808}
}

\end{document}